\documentclass[onecolumn, a4paper,journal,10pt]{IEEEtran}
\usepackage{amsfonts}
\usepackage{graphicx, epsfig,amsmath,amssymb,latexsym,graphics,float}
\usepackage{setspace,subfig}
\usepackage{citesort}
\usepackage{float}
\usepackage[ruled,vlined,linesnumbered]{algorithm2e}
\usepackage{booktabs}
\include{IEEEtran.bib}

\begin{document}

\title{Student's \emph{t} Distribution based Estimation of Distribution Algorithms for Derivative-free Global Optimization}
\author{$^{\ast}$Bin~Liu~\IEEEmembership{Member,~IEEE}, Shi Cheng~\IEEEmembership{Member,~IEEE}, Yuhui Shi~\IEEEmembership{Fellow,~IEEE}
\thanks{B. Liu is with School of Computer Science and Technology, Nanjing University of Posts and Telecommunications, Nanjing,
Jiangsu, 210023 China. S. Cheng is with School of Computer Science, Shaanxi Normal University, Xi'an, 710062 China.
Y. Shi is with Department of Computer Science and Engineering, Southern University of Science and Technology, Shenzhen,
518055 China.

$^{\ast}$ Correspondence author. E-mail: bins@ieee.org}
\thanks{Manuscript received XX. X, 2016; revised XX X, 2016.}}
\maketitle

\begin{abstract}
In this paper, we are concerned with a branch of evolutionary algorithms termed estimation of distribution (EDA),
which has been successfully used to tackle derivative-free global optimization problems.
For existent EDA algorithms, it is a common practice to use a Gaussian distribution or a mixture of Gaussian components to
represent the statistical property of available promising solutions found so far. Observing that the
Student's \emph{t} distribution has heavier and longer tails than the Gaussian, which may be beneficial for
exploring the solution space, we propose a novel EDA algorithm termed ESTDA, in which the Student's \emph{t} distribution, rather than Gaussian,
is employed. To address hard multimodal and
deceptive problems, we extend ESTDA further by substituting a single Student's \emph{t} distribution with a mixture of Student's \emph{t}
distributions. The resulting algorithm is named as estimation of mixture of Student's \emph{t} distribution algorithm (EMSTDA).
Both ESTDA and EMSTDA are evaluated through extensive and in-depth numerical experiments using over a dozen of benchmark objective
functions. Empirical results demonstrate that the proposed algorithms provide remarkably better performance than their Gaussian counterparts.
\end{abstract}

\IEEEkeywords
derivative-free global optimization, EDA, estimation of distribution, Expectation-Maximization, mixture model, Student's \emph{t} distribution

\section{Introduction}\label{sec:intro}
The estimation of distribution algorithm (EDA) is an evolutionary computation (EC) paradigm for tackling derivative-free global optimization
problems \cite{pelikan2002survey}. EDAs have gained great success and attracted increasing attentions during the last decade \cite{hauschild2011introduction}.
Different from most of EC algorithms which are normally equipped with meta-heuristics inspired selection and variation operations,
the EDA paradigm is characterized by its unique variation operation which uses probabilistic models to lead the search towards
promising areas of the solution space. The basic assumption is that probabilistic models can be used for learning useful information of the search
space from a set of solutions that have already been inspected from the problem structure and this information can be used to conduct more
effective search. The EDA paradigm provides a mechanism to take advantage of the correlation structure to drive the search more efficiently.
Henceforth, EDAs are able to provide insights about the structure of the search space.
For most existent EC algorithms, their notions of variation and selection usually arise from the perspective of Darwinian evolution.
These notions offer a direct challenge for quantitative analysis by the existing methods of computer science.
In contrast with such meta-heuristics based EC algorithms, EDAs are more theoretically appealing,
since a wealth of tools and theories arising from statistics or machine learning can be used for EDA analysis.
Convergence results on a class of EDAs have been given in \cite{zhang2004on}.

All existent EDAs fall within the paradigm of model based or learning guided optimization \cite{hauschild2011introduction}. In this regard,
different EDAs can be distinguished by the class of probabilistic models used. An investigation on the boundaries of effectiveness of EDAs
shows that the limits of EDAs are mainly imposed by the probabilistic model they rely on. The more complex the model, the greater ability it
offers to capture possible interactions among the variables of the problem. However, increasing the complexity of the model usually leads to more
computational cost.
The EDA paradigm could admit any type of probabilistic model, among which the most popular model class is the Gaussian model.
The simplest EDA algorithm just employs univariate Gaussian models, which regard all design variables to be independent with each
other \cite{larranaga2002estimation,sebag1998extending}.
The simplicity of models makes such algorithms easy to implement, while their effectiveness halts when the design variables have strong interdependencies.
To get around of this limitation, several multivariate Gaussian based EDAs (Gaussian-EDAs) have been
proposed \cite{larranaga2000optimization,larranaga2002estimation,bosman2000expanding}. To represent variable linkages elaborately,
Bayesian networks (BNs) are usually adopted in
the framework of multivariate Gaussian-EDA, while the learning of the BN structure and parameters can be very time
consuming \cite{dong2013scaling,zhang2004hybrid}. To handle hard multimodal and deceptive problems in a more appropriate way, Gaussian mixture model (GMM)
based EDAs (GMM-EDAs) have also been proposed \cite{lu2005clustering,ahn2008scalability}.
It is worthwhile to mention that the more complex the model used, the more likely it encounters model overfitting, which may mislead the search
procedure \cite{wu2006does}.

The Student's \emph{t} distribution is symmetric and bell-shaped, like the Gaussian distribution, but has heavier and longer tails, meaning that
it is more prone to producing values that fall far from its mean. Its heavier tail property renders the Student's \emph{t} distribution
a desirable choice to design Bayesian simulation techniques such as the adaptive importance sampling (AIS) algorithms \cite{liu2015posterior,liu2014adaptive,cappe2008adaptive}.
Despite the purpose of AIS is totally different as compared with EDAs, AIS also falls within the iterative learning and sampling paradigm like EDAs.
The Student's \emph{t} distribution has also been widely used for robust
statistical modeling \cite{lange1989robust}. Despite of great success achieved in fields of Bayesian simulation and robust modeling,
the Student's \emph{t} distribution has not yet gained any attention from the EC community.

In this paper, we are concerned with whether the heavier tail property of the Student's \emph{t} distribution is beneficial for developing
more efficient EDAs. To this end, we derive two Student's \emph{t} distribution based algorithms within the paradigm of EDA.
They are entitled as estimation of
Student's \emph{t} distribution algorithm (ESTDA) and estimation of mixture of Student's \emph{t} distribution algorithm (EMSTDA), implying
the usages of the Student's \emph{t} distribution and the mixture of Student's \emph{t} distributions, respectively. We evaluate both algorithms
using over a dozen of benchmark objective functions. The empirical results demonstrate that they indeed have better performance than their
Gaussian counterparts. In literature, the closest to our work here are the posterior exploration based Sequential Monte Carlo (PE-SMC) algorithm \cite{liu2015posterior}
and the annealed adaptive mixture importance sampling algorithm \cite{liu2013general}, while both were developed within the framework of Sequential
Monte Carlo (SMC), other than EDA.
In addition, the faster Evolutionary programming (FEP) algorithm proposed in \cite{yao1999evolutionary} is similar in the spirit of
replacing the Gaussian distribution with an alternative that has heavier tails. The major difference between FEP and our work here lies in that
the former falls within the framework of evolutionary strategy while the latter is developed within the EDA paradigm. In addition, the former employs
the Cauchy distribution, other than the student's $t$ or mixture of student's $t$ distributions here, to generate new individuals.
To the best of our knowledge, we make the first attempt here to bring Student's $t$ probabilistic models into the
field of population based EC.

The remainder of this paper is organized as follows. In Section II we revisit the Gaussian-EDAs. Besides the single Gaussian
distribution model based EDA, the GMM-EDA is also presented. In Section III we describe the proposed Student's \emph{t} model
based EDAs in detail. Like in Section II, both the single distribution and the mixture distribution model based EDAs are presented.
In Section IV, we present the performance evaluation results and finally in Section V, we conclude the paper.
\section{Gaussian distribution based EDA}
In this section, we present a brief overview on the EDA algorithmic paradigm, the Gaussian-EDA and the GMM-EDA. 

In this paper, we are concerned with the following derivative-free continuous global optimization problem
\begin{equation}
\underset{x\in\chi}{\min}f(x)
\end{equation}
where $\chi$ denotes the nonempty solution space defined in $\mathbb{R}^n$, and $f$: $\chi\rightarrow\mathbb{R}$ is a continuous real-valued function.
The basic assumption here is that $f$ is bounded on $\chi$, which means $\exists f_l>-\infty, f_u<\infty$ such
that $f_l\leq f(x)\leq f_u$, $\forall x\in\chi$. We denote the minimal function value as $f^{\ast}$, i.e., there exists an $x^{\ast}$ such
that $f(x)\geq f(x^{\ast})=f^{\ast}$, $\forall x\in\chi$.
\subsection{Generic EDA scheme}\label{sec:generic_EDA}
A generic EDA scheme to solve the above optimization problem is presented in Algorithm \ref{algo:EDA}. In this paper, we only
focus on the probabilistic modeling part in the EDA scheme, not intend to discuss the selection operations.
In literature, the tournament or truncation selection are commonly used with the EDA algorithm.
Here we adopt truncation selection for all algorithms under consideration.
Specifically, we select, from $\{x_i\}_{i=1}^N$, $M$ individuals which produce the minimum objective function values.
The stopping criteria of Algorithm \ref{algo:EDA} is specified by the user based on the available budget, corresponding to the
acceptable maximum number of iterations here.
\begin{algorithm}[!htb]
\KwIn{the probabilistic model $p(x|\theta)$ with parameter $\theta$, the population size $N$, the selection size $M$, which
satisfies $M<N$, and $\theta_1$, the initialized value of $\theta$.}
\KwOut{$\hat{x}$ and $f(\hat{x})$, which represent estimates of $x^{\ast}$ and $f(x^{\ast})$, respectively.}
Initialization: Set $\theta=\theta_1$ and then draw a random sample $\hat{x}$ from $p(\cdot|\theta=\theta_1)$. Set the iteration
index $k=1$\;
\While {the stopping criterion is not met} {
    Sample $N$ individuals from $p(x|\theta_k)$, denote them by $x_1,x_2,\ldots,x_N$\;
    Calculate the objective function values of the individuals：$y_i=f(x_i)$, $i=1,2,\ldots,N$ \;
    Find the minimum of $\{y_1,\ldots,y_N\}$, denote it by $y_{min}$ and then find from $\{x_i\}_{i=1}^N$ the corresponding $x_{min}$ which
    produces $y_{min}$\;
    \If{$y_{min}<f(\hat{x})$}{
        Let $\hat{x}=x_{min}$ and $f(\hat{x})=y_{min}$\;
    }
    Select $M$ individuals from $\{x_i\}_{i=1}^N$ based on the information $\{x_i,y_i\}_{i=1}^N$ \;
    Update $\theta_{k-1}$, based on the selected individuals, to get $\theta_k$. See Subsections \ref{sec:gaussian-eda},
    \ref{sec:gmm-eda}, \ref{sec:estda} and \ref{sec:emstda} for details on specific operations with respect to the Gaussian model, GMM,
    the Student's $t$ and the mixture of Student's $t$ models, respectively. 
    Set $k=k+1$.
}
\caption{\label{algo:EDA}A main scheme of EDA}
\end{algorithm}
\subsection{Gaussian-EDA}\label{sec:gaussian-eda}
In Gaussian-EDAs, the probabilistic model $p(x|\theta)$ in Algorithm \ref{algo:EDA} takes the form of Gaussian,
namely $p(x|\theta)=\mathcal{N}(x|\mu,\Sigma)$, where $\theta\triangleq\{\mu,\Sigma\}$, $\mu$ and $\Sigma$ denote the mean and
covariance of this Gaussian distribution, respectively.

As mentioned before, we employ a
truncation selection strategy, which indicates that the selected individuals are those which produce minimum objective function values.
Given the selected individuals $\{x_i,y_i\}_{j=1}^M$, a maximum likelihood (ML) estimate of the parameters of the Gaussian distribution is shown to be:
\begin{eqnarray}
\mu&=&\frac{\sum_{j=1}^Mx_j}{M} \\
\Sigma&=&\frac{\sum_{j=1}^M(x_j-\mu)(x_j-\mu)^T}{M-1},
\end{eqnarray}
where $^T$ denotes transposition and all vectors involved are assumed to be column vectors both here and hereafter.
\subsection{GMM-EDA}\label{sec:GMM-EDA}\label{sec:gmm-eda}
As its name indicates, the GMM-EDA uses a GMM to represent $p(x|\theta)$, namely $p(x|\theta)=\sum_{l=1}^Lw_l\mathcal{N}(x|\mu_l,\Sigma_l)$,
where $\theta\triangleq\{w_l,\mu_l,\Sigma_l\}_{l=1}^L$, $l$ is the index of the mixing components, $w$ the probability mass of the mixing component
which satisfies $\sum_{l=1}^Lw_l=1$, and $L$ denotes the total number of mixing
components included in the mixture and is preset as a constant.

Now we focus on how to update $\theta_{k-1}$ to get $\theta_k$ based on the selected individuals $\{x_j\}_{j=1}^M$.
We use an expectation-maximization (EM) procedure, as shown in Algorithm \ref{algo:EM_GMM}, to update the parameters of the mixture model.
For more details on
EM based parameter estimation for GMM, readers can refer to \cite{bilmes1998gentle,xu1996convergence,biernacki2003choosing,vlassis2002greedy}.
The stopping criterion can be determined by checking the changing rate of the GMM parameters.
If $\theta_{new}$ has not been changed a lot for a fixed number of successive iterations,
then we terminate the iterative process of the EM procedure.
Compared with a benchmark EM procedure, here we add a components deletion operation at the end of the Expectation step. We delete components with extremely
small mixing weights, because their roles in the mixture are negligible, while at the Maximization step,
they will occupy the same computing burden as the rest of components. This deletion operation is also beneficial for avoiding disturbing numerical issues such as singularity of covariance matrix.
For the influence of this operation on the the number of survival mixing components, see Section \ref{sec:experiments_all_alg}.
\begin{algorithm}[!htb]
\KwIn{the individuals $\{x_j\}_{j=1}^M$, the model's current parameter value $\theta_{k-1}=\{w_{k-1,l},\mu_{k-1,l},\Sigma_{k-1,l}\}_{l=1}^L$
and the admittable smallest weight $W$ of a mixing component (we set $W=0.02$ as a default value for use in our experiments)}
\KwOut{$\theta_{k}$}
Initialization: Set $\theta_{old}=\theta_{k-1}$. Set the iteration index $i=1$\;
\While {the stopping criterion is not met} {
    The Expectation step:
    \begin{equation}
     w_{new,l}=\sum_{j=1}^M\epsilon(l|x_j)/M,\quad l=1,\ldots, L,
     \end{equation}
     where $\epsilon(l|x_j)$ denotes the probability of the event individual $x_j$ belonging to the $l$th mixing component, which is calculated as follows
     \begin{equation}
      \epsilon(l|x_j)=\frac{w_{old,l}\mathcal{N}(x_j|\mu_{old,l},\Sigma_{old,l})}{\sum_{l=1}^Lw_{old,l}\mathcal{N}(x_j|\mu_{old,l},\Sigma_{old,l})}.
     \end{equation}
     Delete mixing components whose mixing weights are smaller than $W$. Then update the value of $L$ and increase the mixing weights of the remaining mixing
     components proportionally to guarantee that their summation equals 1\;
     The Maximization step:
     \begin{eqnarray}
     \mu_{new,l}&=&\sum_{j=1}^M\epsilon(l|x_j)x_j/M,\quad l=1,\ldots, L,\\
     \Sigma_{new,l}&=&\sum_{j=1}^M\epsilon(l|x_j)(x_j-\mu_{new,l})(x_j-\mu_{new,l})^T/M,\quad l=1,\ldots, L.
     \end{eqnarray}
     Set $\theta_{new}=\{w_{new,l},\mu_{new,l},\Sigma_{new,l}\}_{l=1}^L$\;
     Set $\theta_{old}=\theta_{new}$ and let $k=k+1$.
}
Set $\theta_{k}=\theta_{new}$\;
\caption{\label{algo:EM_GMM}EM procedure to estimate parameters of the GMM within the EDA paradigm as presented in Algorithm 1}
\end{algorithm}
\section{Student's t distribution based EDA}
In this section, we describe the proposed ESTDA and EMSTDA in detail.
Both of them are derived based on the application of the Student's \emph{t} distribution in the EDA framework as presented in Algorithm \ref{algo:EDA}.
To begin with, we give a brief introduction to the Student's \emph{t} distribution.
\subsection{Student's \emph{t} distribution}\label{sec:t}
\begin{figure}[htb]
\begin{tabular}{c}
\centerline{\includegraphics[width=6in,height=4in]{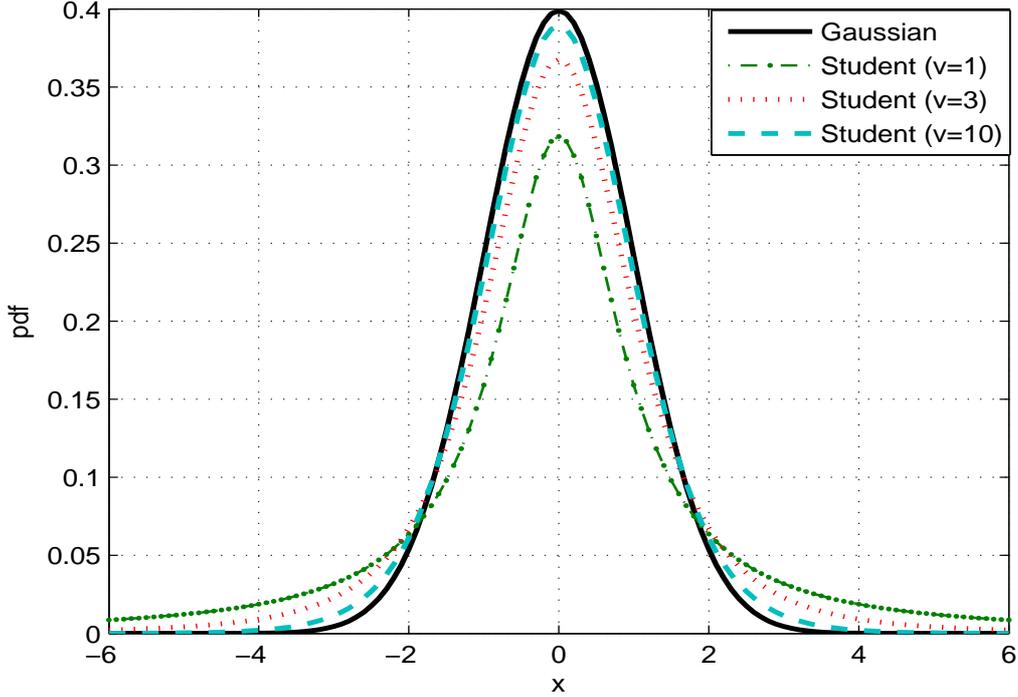}}
\end{tabular}
\caption{Comparison between the univariate Gaussian and Student's \emph{t} distributions (represented by ``Student" in the figure).
All distributions have zero mean with a fixed variance 1. $v$ denotes the DoF.}\label{fig:compare_t_normal}
\end{figure}

Suppose that $x$ is a $d$ dimensional random variable that follows the multivariate Student's $t$ distribution,
denoted by $\mathcal{S}(\cdot|\mu,\Sigma,v)$, where $\mu$ denotes the mean, $\Sigma$ a positive definite inner product matrix and $v\in(0,\infty]$ is
the degrees of freedom (DoF). Then the density function of $x$ is:
\begin{equation}\label{def_t}
\mathcal{S}(x|\mu,\Sigma,v)=\frac{\Gamma(\frac{v+d}{2})|\Sigma|^{-0.5}}{(\pi v)^{0.5d}\Gamma(\frac{v}{2})\{1+M_d(x,\mu,\Sigma)/v\}^{0.5(v+d)}},
\end{equation}
where
\begin{equation}
M_d(x,\mu,\Sigma)=(x-\mu)^T \Sigma^{-1}(x-\mu)
\end{equation}
denotes the Mahalanobis squared distance from $x$ to $\mu$ with
respect to $\Sigma$, $A^{-1}$ denotes the inverse of $A$ and $\Gamma(\cdot)$ denotes the gamma function. For ease of understanding, we give a graphical description of univariate Student's $t$ pdfs
corresponding to different DoFs in comparison with a standard Gaussian pdf in Fig.\ref{fig:compare_t_normal}. It is shown that the
Student's $t$ pdfs have heavier
tails than the Gaussian pdf. The smaller the DoF $v$, the heavier the corresponding pdf's tails.
From Fig.\ref{fig:compare_t_normal}, we see that the Student's $t$ pdf is more likely
to generate an individual further away from its mean than
the Gaussian model due to its long flat tails. It leads to a higher probability of escaping from a local optimum
or moving away from a plateau.
We can also see that heavier tails lead to a smaller hill around the mean, and vice versa.
A smaller hill around the mean indicates that the corresponding model tuning operation spends less time in exploiting the
local neighborhood and thus has a weaker fine-tuning ability. Hence heavier tails do not always bring advantages. This has been demonstrated by
our numerical tests. See Fig.\ref{fig:t_compare_subfig_d}, which shows that for several two-dimensional (2D) benchmark test problems, lighter tails associated with $v=50$ provide better convergence result than
heavier tails corresponding to $v=5$ and 10.
\subsection{ESTDA}\label{sec:estda}
In ESTDA, the probabilistic model $p(x|\theta)$ in Algorithm \ref{algo:EDA} is specified to be a student's $t$ distribution,
namely $p(x|\theta)=\mathcal{S}(x|\mu,\Sigma,v)$, where $\theta\triangleq\{\mu,\Sigma\}$, and $v$ is specified beforehand as a constant.

First, let's figure out, given the parameter value $\theta$, how to sample an individual $x$ from $p(\cdot|\theta)$.
It consists of two steps that is simulating a random draw $\tau$ from the gamma distribution, whose shape and
scale parameters are set identically to $v/2$, and sampling $x$ from the Gaussian distribution $\mathcal{N}(\cdot|\mu,\Sigma/\tau)$.

Now let's focus on, given a set of selected individuals $\{x_j\}_{j=1}^M$, how to update the
parameters of the student's $t$ distribution in a optimal manner in terms of maximizing likelihood. As mentioned above, in generation of $\{x_j\}_{j=1}^M$,
a corresponding set of gamma distributed variables $\{\tau_j\}_{j=1}^M$ is used. We can record these gamma variables and then use them to easily
derive an ML estimate for parameters of the Student's $t$ distribution \cite{liu1997ml,liu1995ml}:
\begin{eqnarray}
\mu&=&\frac{\sum_{j=1}^M \tau_jx_j}{\sum_{i=1}^M\tau_i} \\
\Sigma&=&\frac{\sum_{j=1}^M \tau_j(x_j-\mu)(x_j-\mu)^T}{\sum_{i=1}^M\tau_i}.
\end{eqnarray}
\subsection{EMSTDA}\label{sec:emstda}
The EMSTDA employs a mixture of student's $t$ distributions to play the role of the probabilistic model $p(x|\theta)$ in Algorithm \ref{algo:EDA}.
Now we have $p(x|\theta)=\sum_{l=1}^Lw_l\mathcal{S}(x|\mu_l,\Sigma_l,v)$, where $\theta\triangleq\{w_l,\mu_l,\Sigma_l\}_{l=1}^L$, $\sum_{l=1}^Lw_l=1$,
and $v$ is specified beforehand as a constant. Given the selected individuals $\{x_j\}_{j=1}^M$, we resort to the EM method to update parameter
values of the mixture of Student's $t$ distributions. The specific operations are presented in Algorithm \ref{algo:EM_t_mixture}.
EM methods based parameter estimation for the mixture of Student's $t$ model can also be found in \cite{liu2014adaptive,liu2015posterior,cappe2008adaptive},
where the mixture learning procedure is performed based on a weighted sample set. The EM operations presented in Algorithm \ref{algo:EM_t_mixture}
can be regarded as an application of the EM methods described in \cite{liu2014adaptive,liu2015posterior,cappe2008adaptive} to an equally weighted sample set.
Similarly as in Algorithm \ref{algo:EM_GMM}, we add a components deletion operation at the end of Expectation step.
This deletion operation were used in the adaptive annealed importance sampling \cite{liu2014adaptive} algorithm and the PE-SMC algorithm \cite{liu2015posterior},
which also adopt the EM procedure to do parameter estimation for a Student's $t$ mixture model.
By deleting components with extremely small mixing weights, we can avoid consuming computing burdens to update parameter values for negligible components.
As presented before in Subsection \ref{sec:GMM-EDA}, this deletion operation is also useful to avoid common
numerical issues such as singularity of covariance matrix. For the influence of this operation on the the number of survival mixing components,
see Subsection \ref{sec:experiments_all_alg}.
\begin{algorithm}[!htb]
\KwIn{the individuals $\{x_j\}_{j=1}^M$, the model's current parameter value $\theta_{k-1}=\{w_{k-1,l},\mu_{k-1,l},\Sigma_{k-1,l}\}_{l=1}^L$
and the admittable smallest weight $W$ of a mixing component (we set $W=0.02$ as a default value for use in our experiments)}
\KwOut{$\theta_{k}$}
Initialization: Set $\theta_{old}=\theta_{k-1}$. Set the iteration index $i=1$\;
\While {the stopping criterion is not met} {
    The Expectation step:
    \begin{equation}
     w_{new,l}=\sum_{j=1}^M\epsilon(l|x_j)/M,\quad l=1,\ldots, L,
     \end{equation}
     where $\epsilon(l|x_j)$ denotes the probability of the event individual $x_j$ belonging to the $l$th mixing component, which is calculated as follows
     \begin{equation}
      \epsilon(l|x_j)=\frac{w_{old,l}\mathcal{S}(x_j|\mu_{old,l},\Sigma_{old,l},v)}{\sum_{l=1}^Lw_{old,l}\mathcal{S}(x_j|\mu_{old,l},\Sigma_{old,l},v)}.
     \end{equation}
     Delete mixing components whose mixing weights are smaller than $W$. Then update the value of $L$ and increase the mixing weights of the remaining mixing
     components proportionally to guarantee that their summation equals 1\;
     The Maximization step:
     \begin{eqnarray}
     \mu_{new,l}&=&\frac{\sum_{j=1}^M\epsilon(l|x_j)x_j(v+d)/\left(v+M_d(x_j,\mu_{old,l},\Sigma_{old,l})\right)}{\sum_{j=1}^M\epsilon(l|x_j)(v+d)/\left(v+M_d(x_j,\mu_{old,l},\Sigma_{old,l})\right)},\; l=1,\ldots, L,\\
     \Sigma_{new,l}&=&\frac{\sum_{j=1}^M\epsilon(l|x_j)(x_j-\mu_{new,l})(x_j-\mu_{new,l})^T(v+d)/\left(v+M_d(x_j,\mu_{old,l},\Sigma_{old,l})\right)}{\sum_{j=1}^M\epsilon(l|x_j)},\; l=1,\ldots, L.
     \end{eqnarray}
     Set $\theta_{new}=\{w_{new,l},\mu_{new,l},\Sigma_{new,l}\}_{l=1}^L$\;
     Set $\theta_{old}=\theta_{new}$ and let $k=k+1$.
}
Set $\theta_{k}=\theta_{new}$\;
\caption{\label{algo:EM_t_mixture}EM procedure to estimate parameters of a mixture of student's $t$ model within the EDA paradigm as presented in Algorithm 1}
\end{algorithm}
\section{Performance evaluation}\label{sec:performance_compare}
In this section, we evaluate the presented ESTDA and EMSTDA using a number of benchmark objective functions, which are
designed and commonly used in literature for
testing and comparing optimization algorithms. See the Appendix Section for details on the involved functions.
The Gaussian-EDA and the GMM-EDA, which are described in Sections \ref{sec:gaussian-eda} and \ref{sec:gmm-eda}, respectively, are involved for comparison purpose.

Although scaling up EDAs to large scale problems has become one of the biggest challenges of the field \cite{Kaban2015},
it is not our intention here to discuss and investigate the applications of ESTDA and EMSTDA in high dimensional problems.
The goal of this paper is to demonstrate the benefits resulted from the heavier tails of Student's $t$ distribution
in exploring the solution space and finding the global optimum through the EDA mechanism.
Therefore we only consider cases with $d\leq10$ here for ease of performance comparison.
\subsection{Experimental study for ESTDA}\label{sec:experiment_estda}
As described in Subsection \ref{sec:t}, the degree of difference between the shapes of a Student's $t$ and a Gaussian distribution mainly depends on
the value of DoF $v$. This Subsection is dedicated to investigating the influence of $v$ on the
performance of ESTDA through empirical studies.

\begin{figure}[htbp]
\centering
\subfloat[Ackley Function]{
\label{fig:t_compare_subfig_a}
\begin{minipage}[t]{0.3\textwidth}
\includegraphics[width=2.2in,height=1.3in]{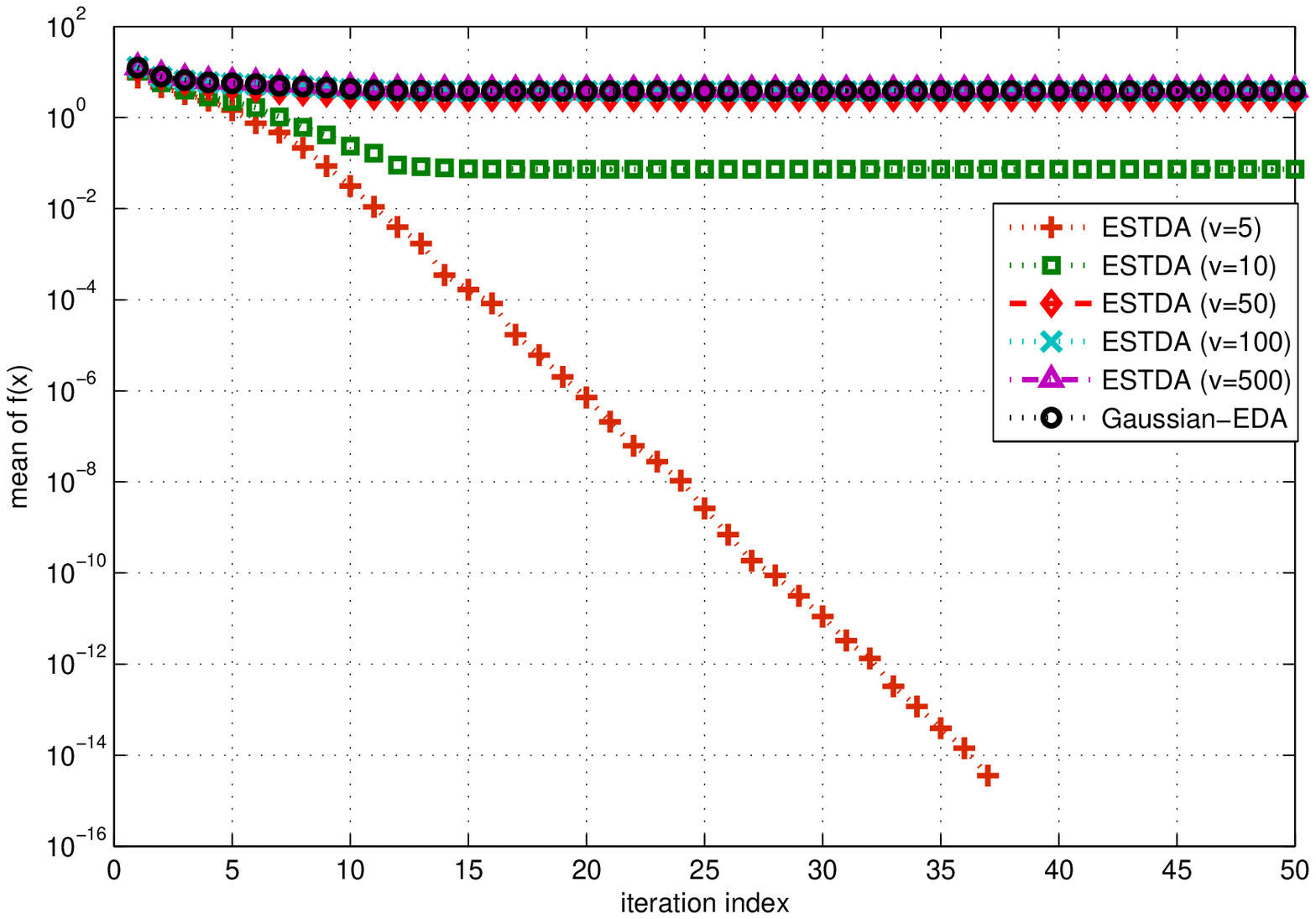}
\end{minipage}
}
\subfloat[De Jong Function N.5]{
\label{fig:t_compare_subfig_b}
\begin{minipage}[t]{0.3\textwidth}
\includegraphics[width=2.2in,height=1.3in]{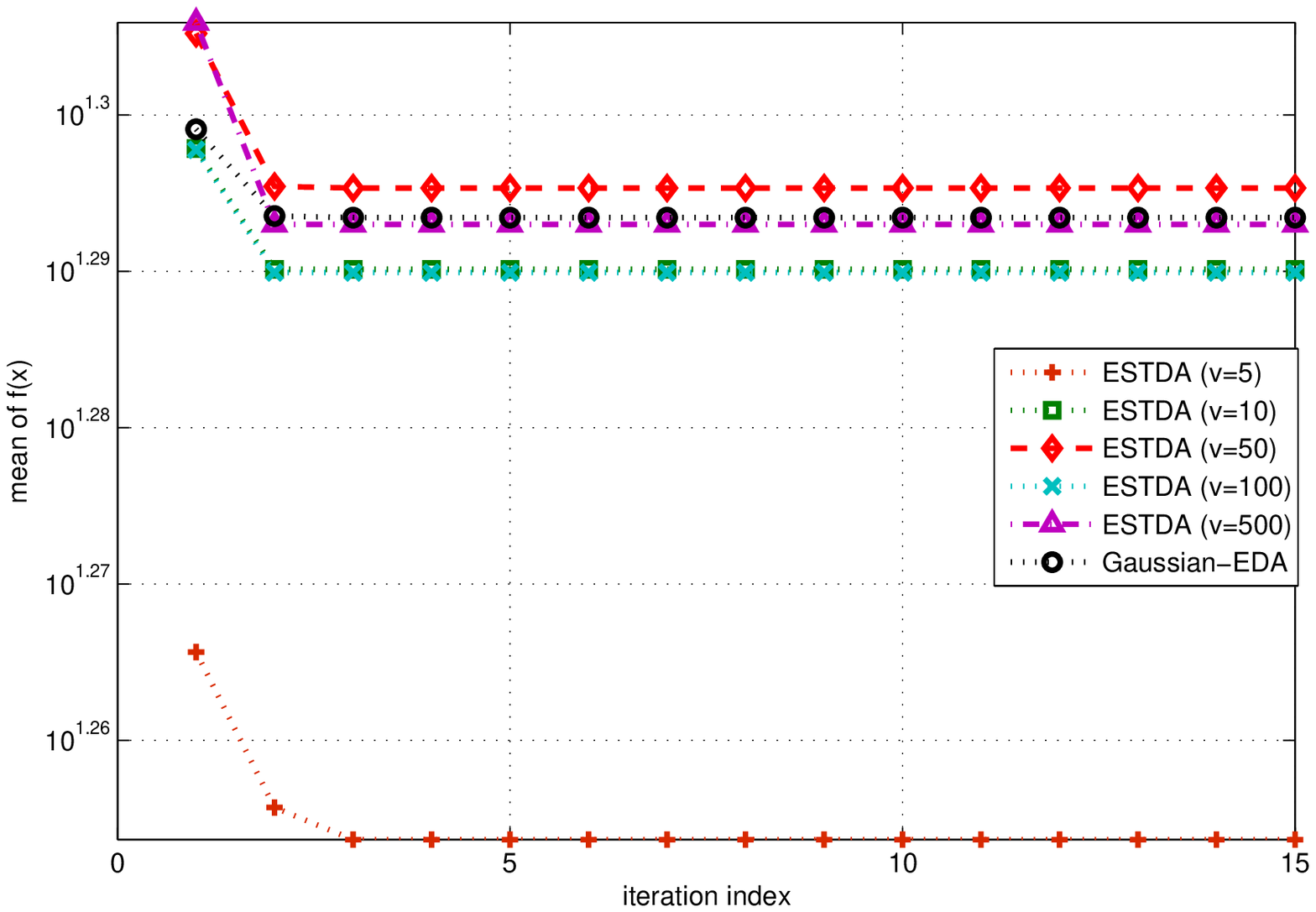}
\end{minipage}
}
\subfloat[Easom Function]{
\label{fig:t_compare_subfig_c}
\begin{minipage}[t]{0.3\textwidth}
\includegraphics[width=2.2in,height=1.3in]{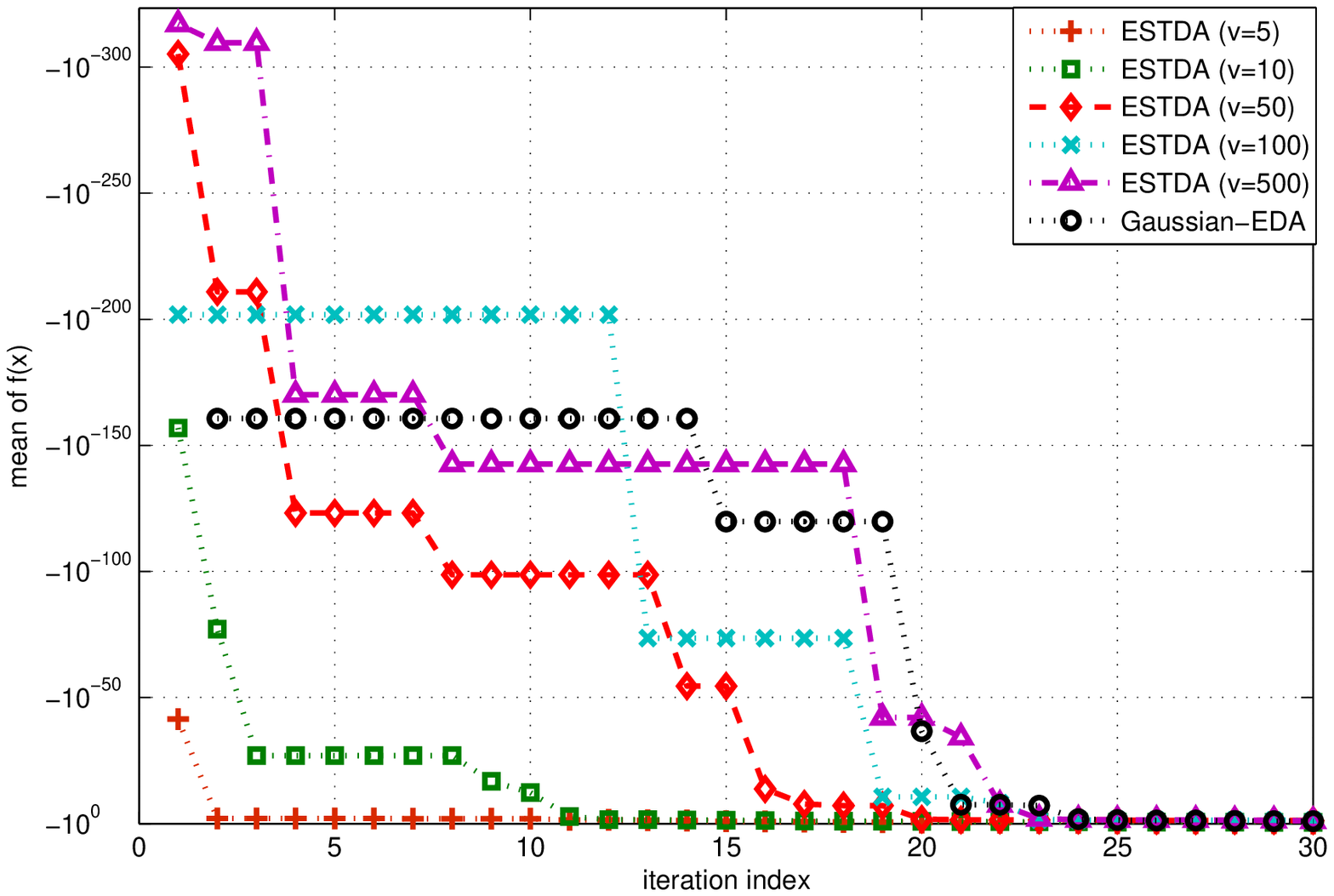}
\end{minipage}
}\\
\subfloat[Rastrigin Function]{
\label{fig:t_compare_subfig_d}
\begin{minipage}[t]{0.3\textwidth}
\includegraphics[width=2.2in,height=1.3in]{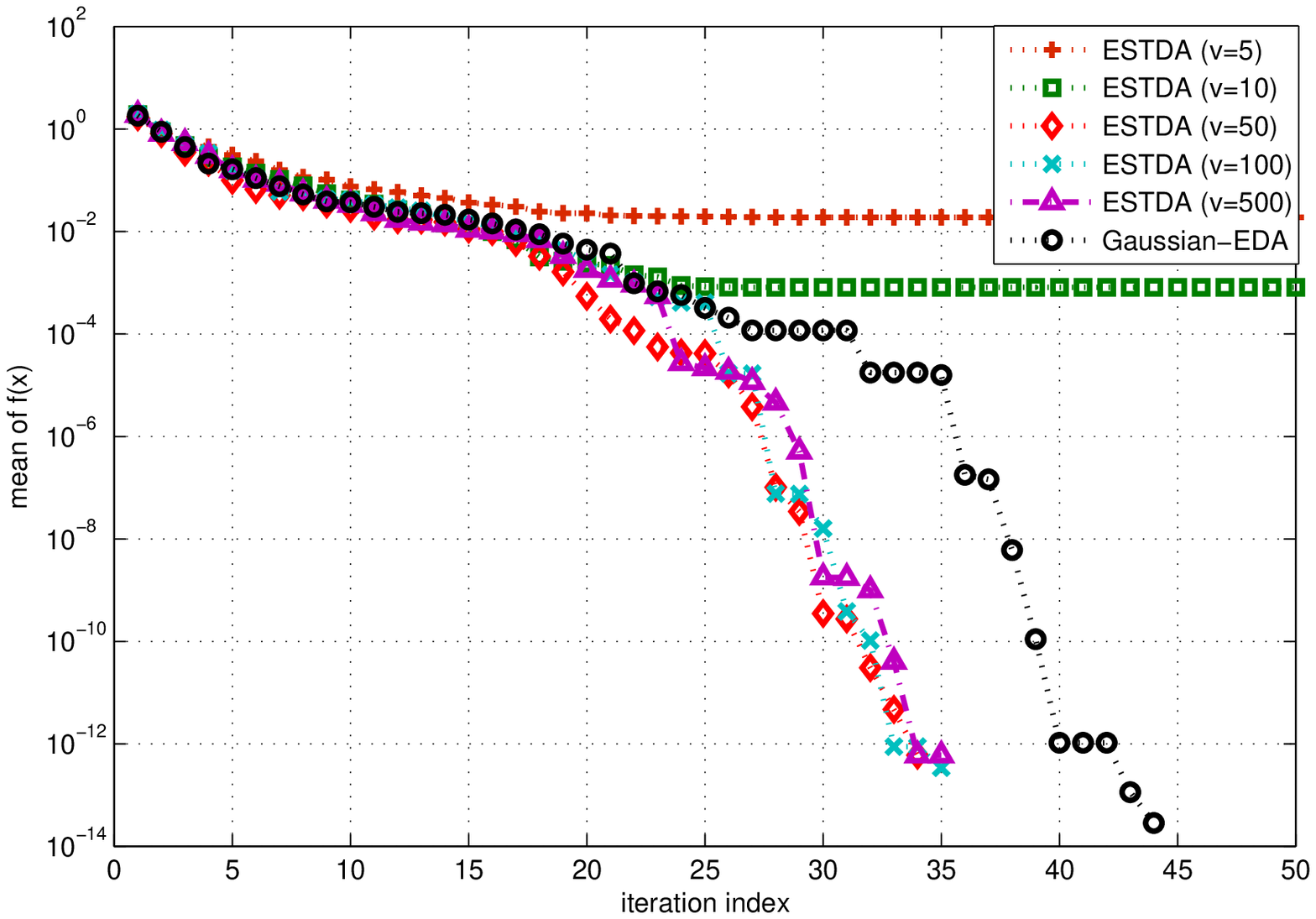}
\end{minipage}
}
\subfloat[Michalewicz Function]{
\label{fig:t_compare_subfig_e}
\begin{minipage}[t]{0.3\textwidth}
\includegraphics[width=2.2in,height=1.3in]{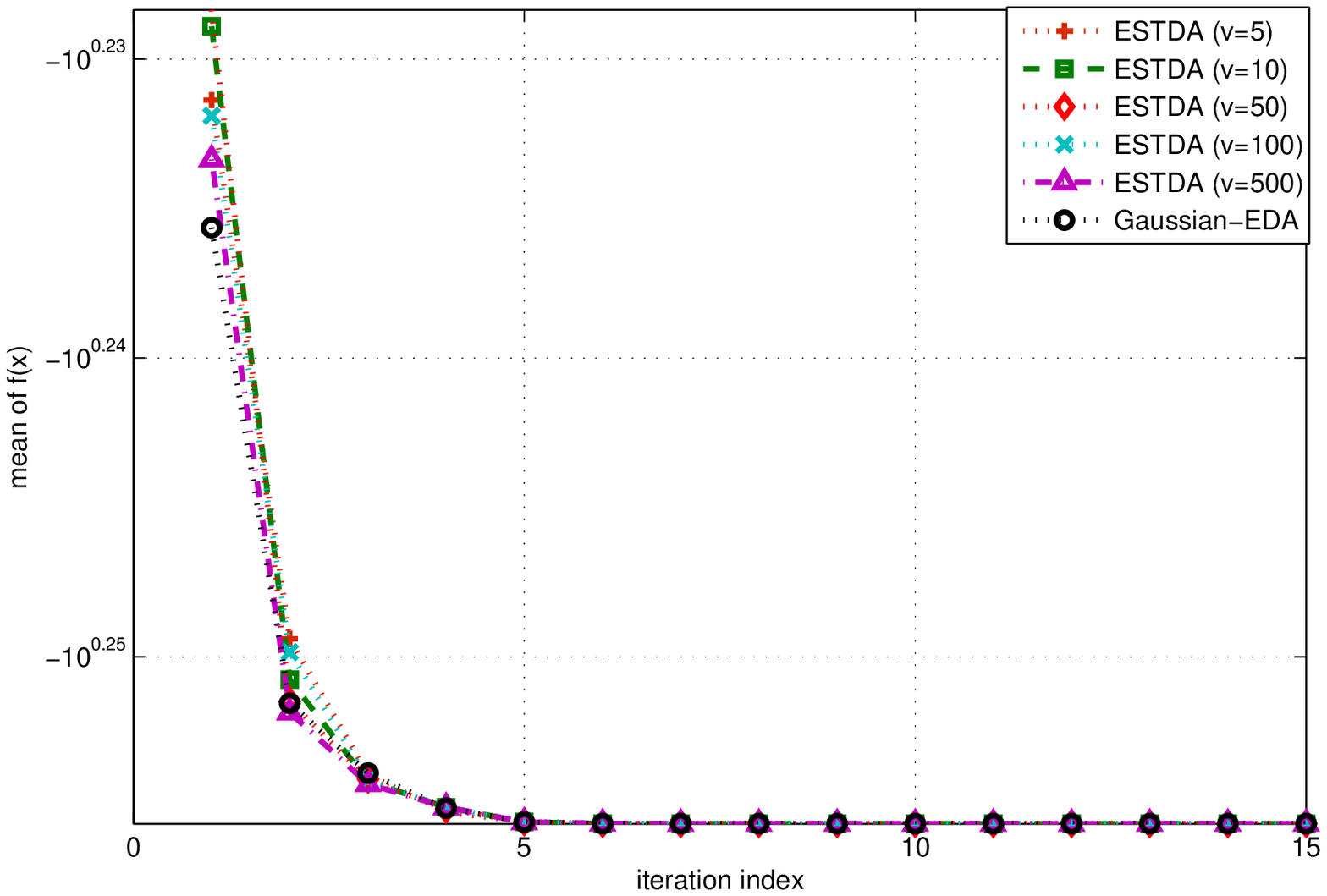}
\end{minipage}
}
\subfloat[L\`{e}vy N.13 Function]{
\label{fig:t_compare_subfig_f}
\begin{minipage}[t]{0.3\textwidth}
\includegraphics[width=2.2in,height=1.3in]{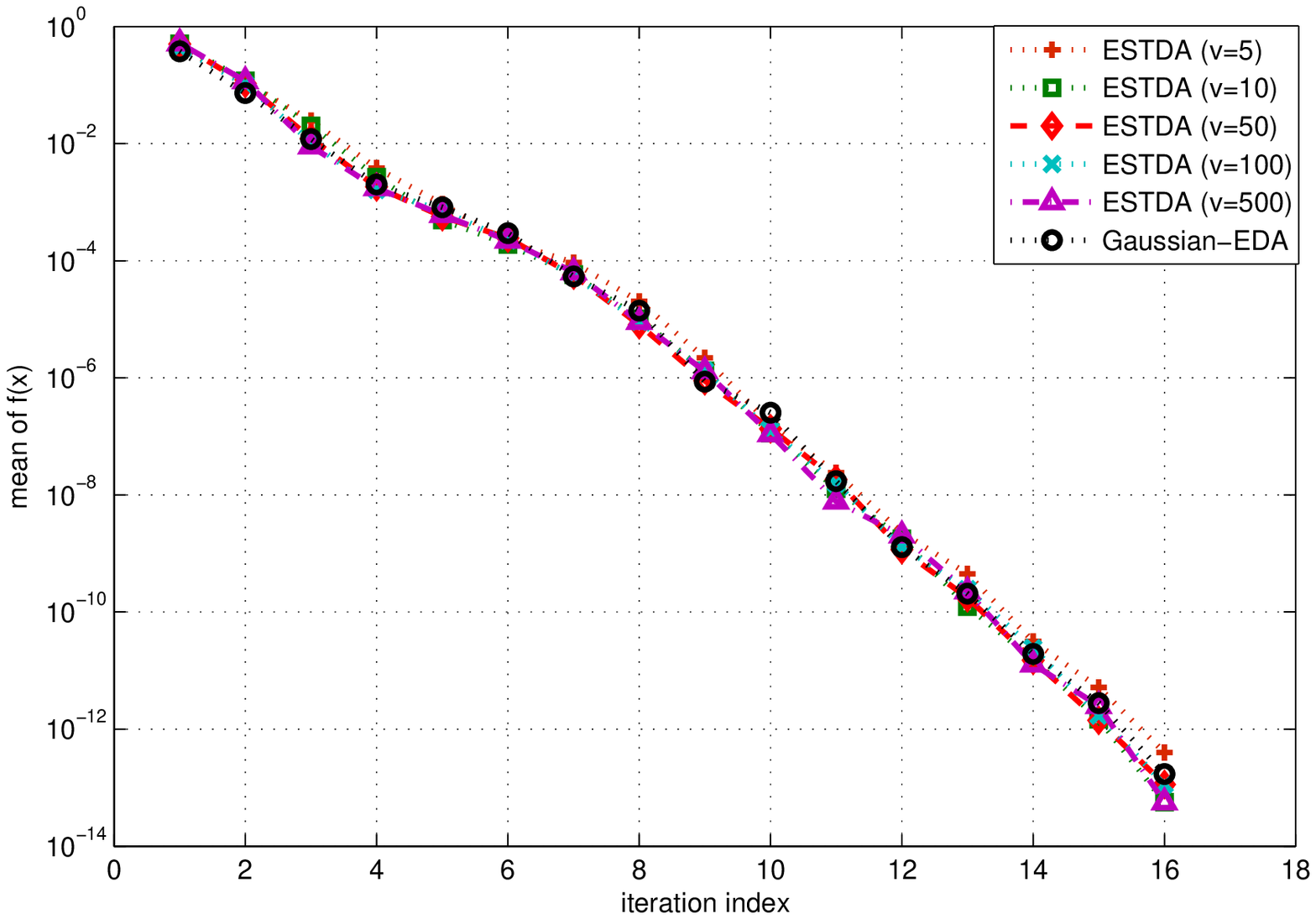}
\end{minipage}
}
\caption{Convergence behaviors of ESTDAs corresponding to different DoFs and a Gaussian-EDA.
The Y coordinate denotes the averaged objective function values of the best solutions obtained from 30
independent runs of each algorithm. All test functions involved here are 2D}\label{fig:t_compare}
\end{figure}

We will test on six objective functions, including the Ackley, De Jong Function N.5, Easom, Rastrigin,
Michalewicz and the L\`{e}vy N.13 functions. We consider the 2D cases (i.e., $d$=2) here. These functions have diversing characteristics in their shapes and thus give representativeness
to the results we will report. See the Appendix for mathematical definitions of these functions.
We consider four ESTDAs, corresponding to different DoFs $v=5, 10, 50, 100, 500$, respectively.
The sample size and selection size are fixed to be $N=1000$ and $M=200$ for all ESTDAs under consideration.
A basic Gaussian-EDA is involved, acting as a benchmark algorithm for comparison.
We run each algorithm 30 times independently and record the best solutions it obtained at the end of each iteration.
We plot the means of the best fitness function values obtained over these 30 runs in Fig.\ref{fig:t_compare}.
It is shown that the ESTDA $(v=5)$, performs strikingly better than the other candidate algorithms for the Ackley, De Jong Function N.5 and Easom Functions.
The ESTDA $(v=50)$ and ESTDA $(v=500)$ outperform the others significantly in handling the Rastrigin function; and, for the remaining
Michalewicz and L\`{e}vy N.13 functions, all the algorithms converge to the optima at a very similar rate.
To summarize, the ESTDA outperforms the Gaussian-EDA markedly for four problems and performs similarly as the Gaussian-EDA for
the other two problems considered here.
This result demonstrate that, the heavier tails of the Student's $t$ distribution can facilitate the process of
exploring the solution space and thus constitute a beneficial factor for the EDA algorithms to find the global optimum faster.
\subsection{A more thorough performance evaluation for ESTDA and EMSTDA}\label{sec:experiments_all_alg}
In this subsection, we perform a more thorough performance evaluation for ESTDA and EMSTDA.
Both Gassuian-EDA and GMM-EDA are also involved for comparison purpose. Including GMM-EDA we hope to further demonstrate that the
heavier tails of the Student's $t$ distribution are beneficial for exploration of solution space via the EDA mechanism.
By contrasting EMSTDA with ESTDA, we can inspect how much a mixture model can contribute to search the solution space via a
Student's $t$ based EDA mechanism.

To make a representative evaluation, we will consider in total 17 objective functions listed in Table \ref{Table:convergence values}.
For all but the Rastrigin and Michalewicz functions, we consider 2D cases. For the Rastrigin and Michalewicz functions, we also consider
5D and 10D cases. The population size per iteration $N$ is set to be 1e3, 1e4, and 1e5 for 2D, 5D, 10D cases, respectively.
The selection size $M$ is fixed to be $0.2\times N$ for all cases involved.
We didn't consider higher dimensional cases, for which a reliable probabilistic model building requires far more samples,
while it is impractical to carry out so large numbers of samples in reasonable
time scales. Definitions of all the involved objective functions are presented in the Appendix of the paper.

We fix the value of the DoF $v$ to be 5 for all test functions, except that, for problems involving the Rastrigin function, $v$ is set to be 50,
since it gives better convergence behavior as reported in
subsection \ref{sec:experiment_estda}.
For every algorithm, the iterative EDA procedure terminates when the iteration number $k$ in Algorithm \ref{algo:EDA} is bigger than 50.
For EMSTDA and GMM-EDA, the iterative EM procedure terminates when the iteration number $i$ in Algorithms \ref{algo:EM_GMM}
and \ref{algo:EM_t_mixture} is bigger than 2.
Each algorithm is run 30 times independently. Then we calculate the average and standard error of the converged fitness values.
The results are given in Table \ref{Table:convergence values}. For each test function, the best solution is marked with bold font.
Then we count for each algorithm how many times it output a better solution as compared with all the rest of algorithms and put the result into Table
\ref{Table:stat}. We see that the EMSTDA get the highest score 7, followed by score 5 obtained by ESTDA, while their Gaussian counterparts
only obtain relatively lower scores 0 and 2, respectively. This result further coincides with our argument that the Student's $t$
distribution is more preferable than the Gaussian for use in designing EDAs.
We also see that, between ESTDA and EMSTDA, the latter performs better than the former. This demonstrates that using mixture models
within a Student's $t$ based EDA is likely to be able to bring additional advantages.
What slightly surprises us in Table \ref{Table:stat} is that the GMM-EDA seems to perform worse than the Gaussian-EDA, while after
a careful inspection on Table \ref{Table:convergence values}, we see that GMM-EDA beats Gaussian-EDA strikingly in cases involving
the Ackley, Dejong N.5, Easom, Michalewicz 10D, Eggholder, Griewank, Holder table, Schwefel and Rosenbrock functions,
while for cases involving the Michalewicz 2D, L\`{e}vy N.13, Cross-in-tray, Drop-wave, L\`{e}vy, Schaffer, Shubert and Perm functions,
GMM-EDA performs identically or similarly as the Gaussian-EDA. Hence, in summarize, GMM-EDA actually performs better than Gaussian-EDA in this
experiment.

\begin{figure}[htbp]
\centering
\subfloat[Ackley Function]{
\label{fig:L_compare_subfig_a}
\begin{minipage}[t]{0.3\textwidth}
\includegraphics[width=2.2in,height=1.3in]{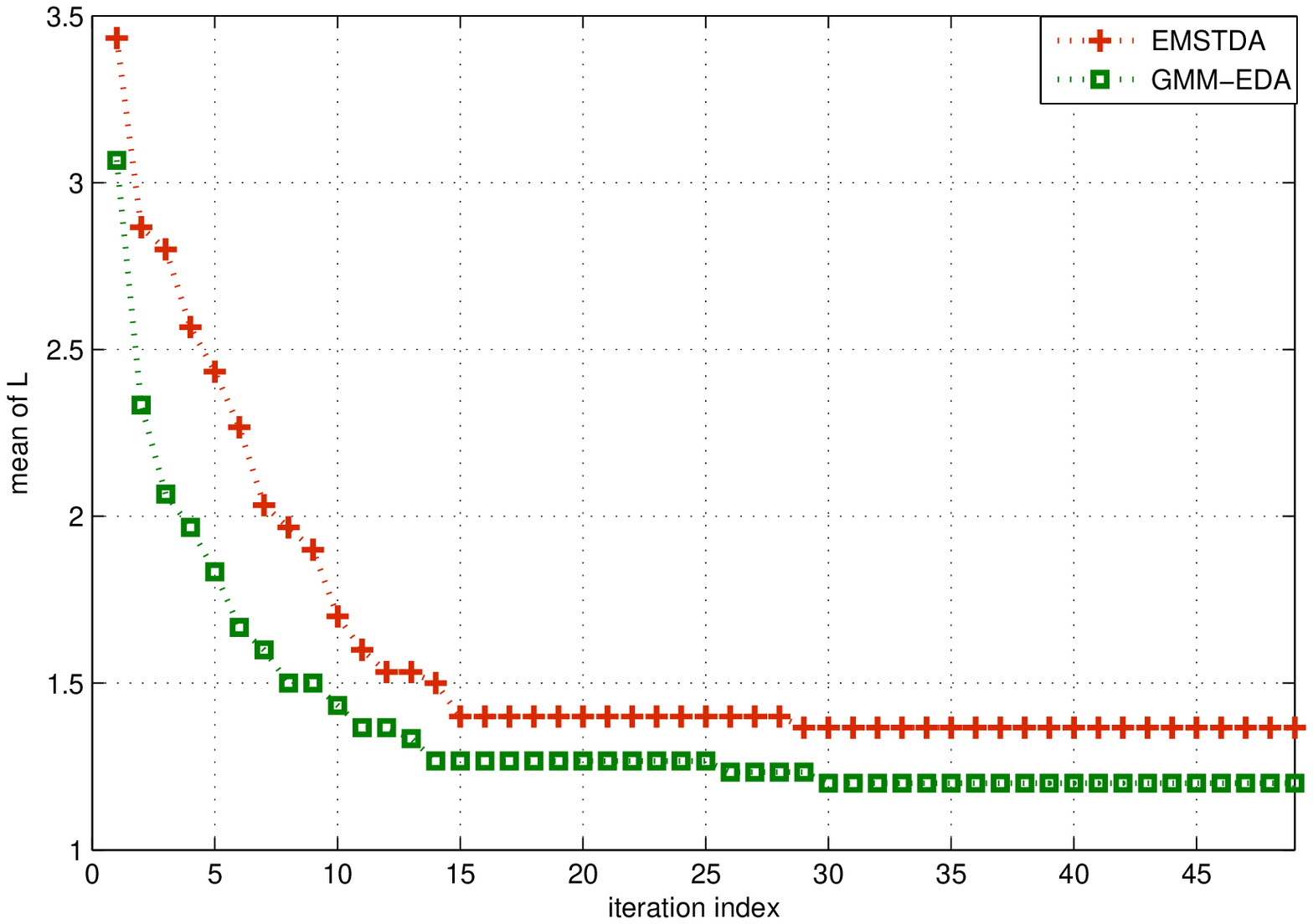}
\end{minipage}
}
\subfloat[De Jong Function N.5]{
\label{fig:L_compare_subfig_b}
\begin{minipage}[t]{0.3\textwidth}
\includegraphics[width=2.2in,height=1.3in]{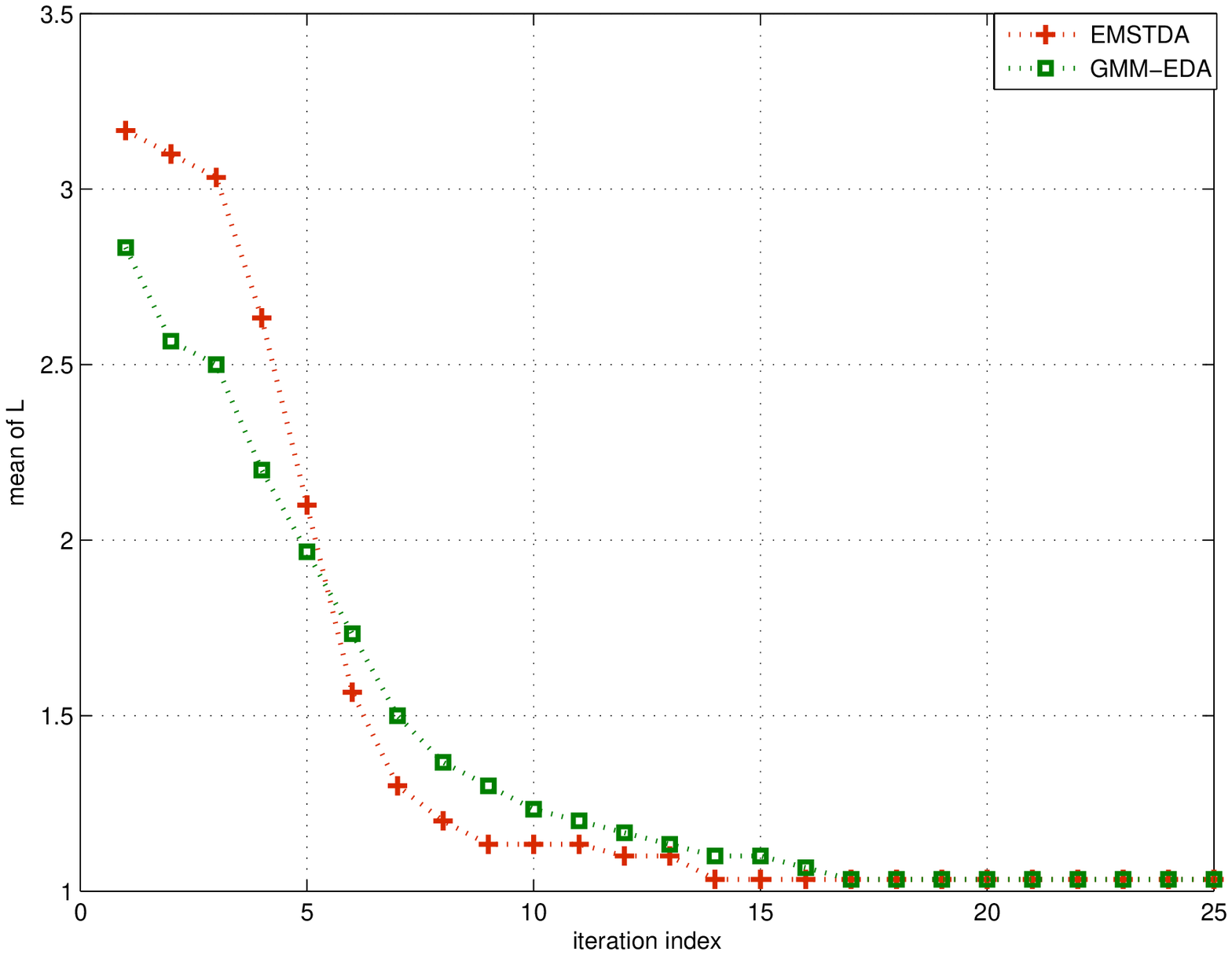}
\end{minipage}
}
\subfloat[Easom Function]{
\label{fig:L_compare_subfig_c}
\begin{minipage}[t]{0.3\textwidth}
\includegraphics[width=2.2in,height=1.3in]{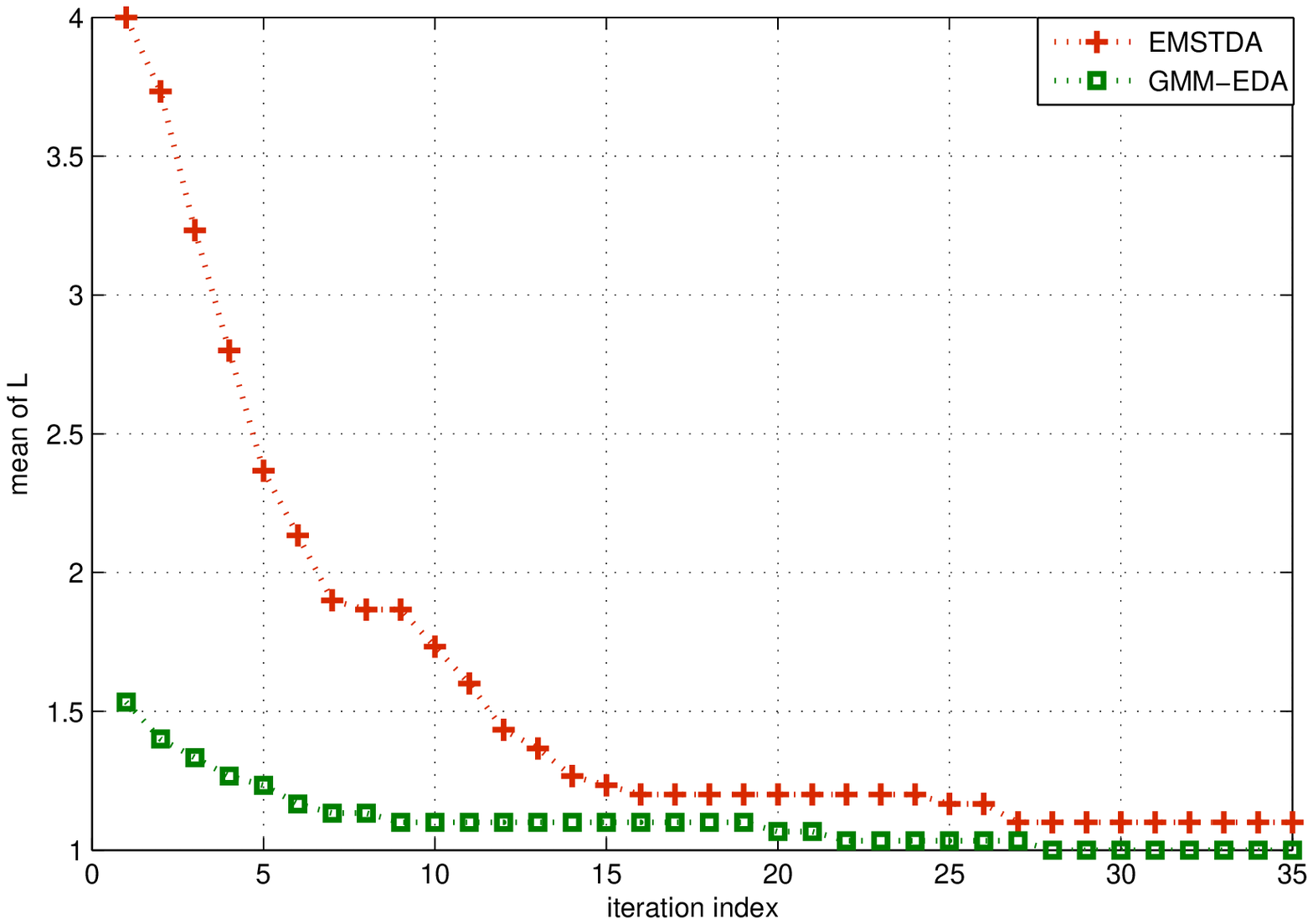}
\end{minipage}
}\\
\subfloat[Rastrigin Function]{
\label{fig:L_compare_subfig_d}
\begin{minipage}[t]{0.3\textwidth}
\includegraphics[width=2.2in,height=1.3in]{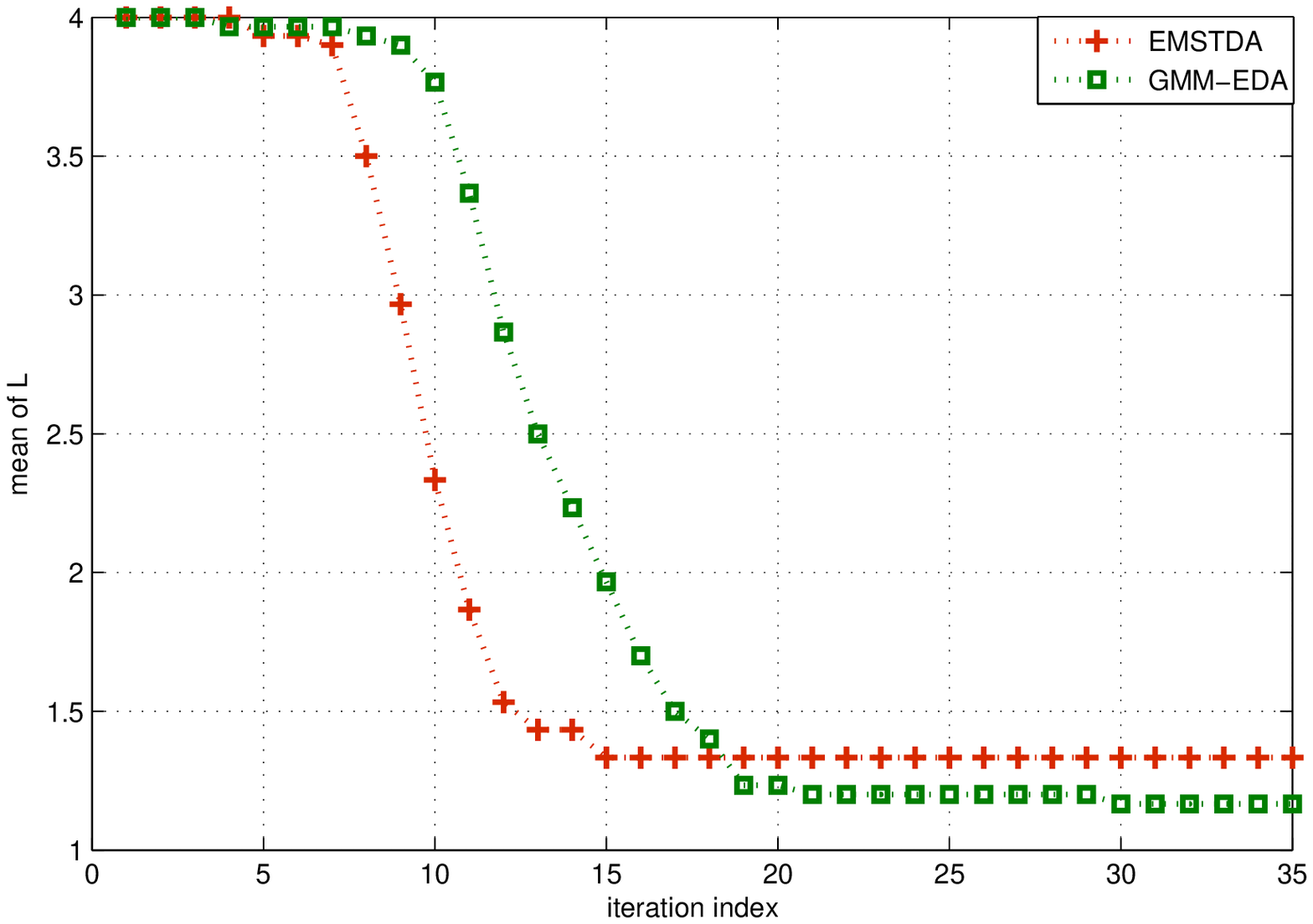}
\end{minipage}
}
\subfloat[Michalewicz Function]{
\label{fig:L_compare_subfig_e}
\begin{minipage}[t]{0.3\textwidth}
\includegraphics[width=2.2in,height=1.3in]{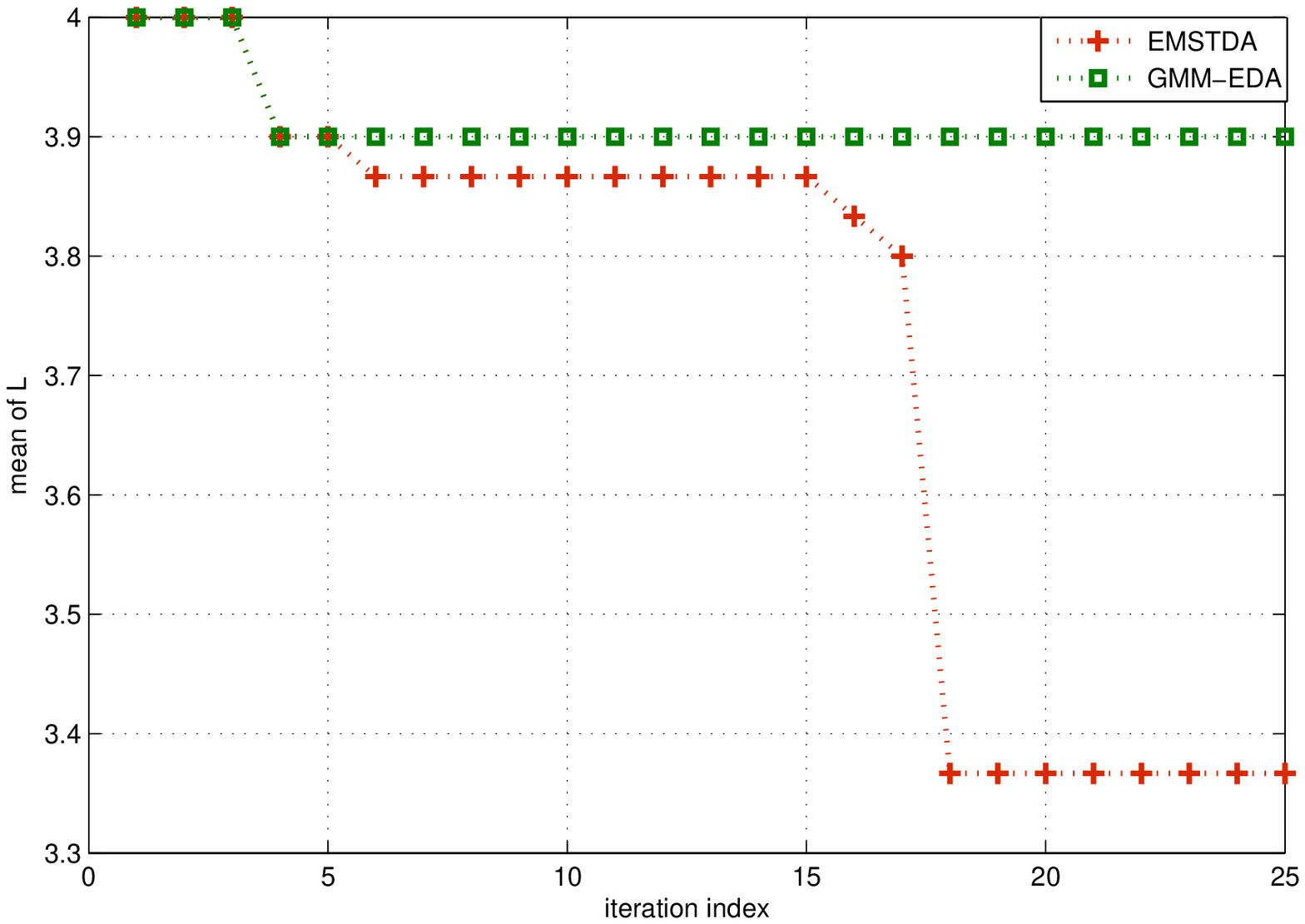}
\end{minipage}
}
\subfloat[L\`{e}vy N.13 Function]{
\label{fig:L_compare_subfig_f}
\begin{minipage}[t]{0.3\textwidth}
\includegraphics[width=2.2in,height=1.3in]{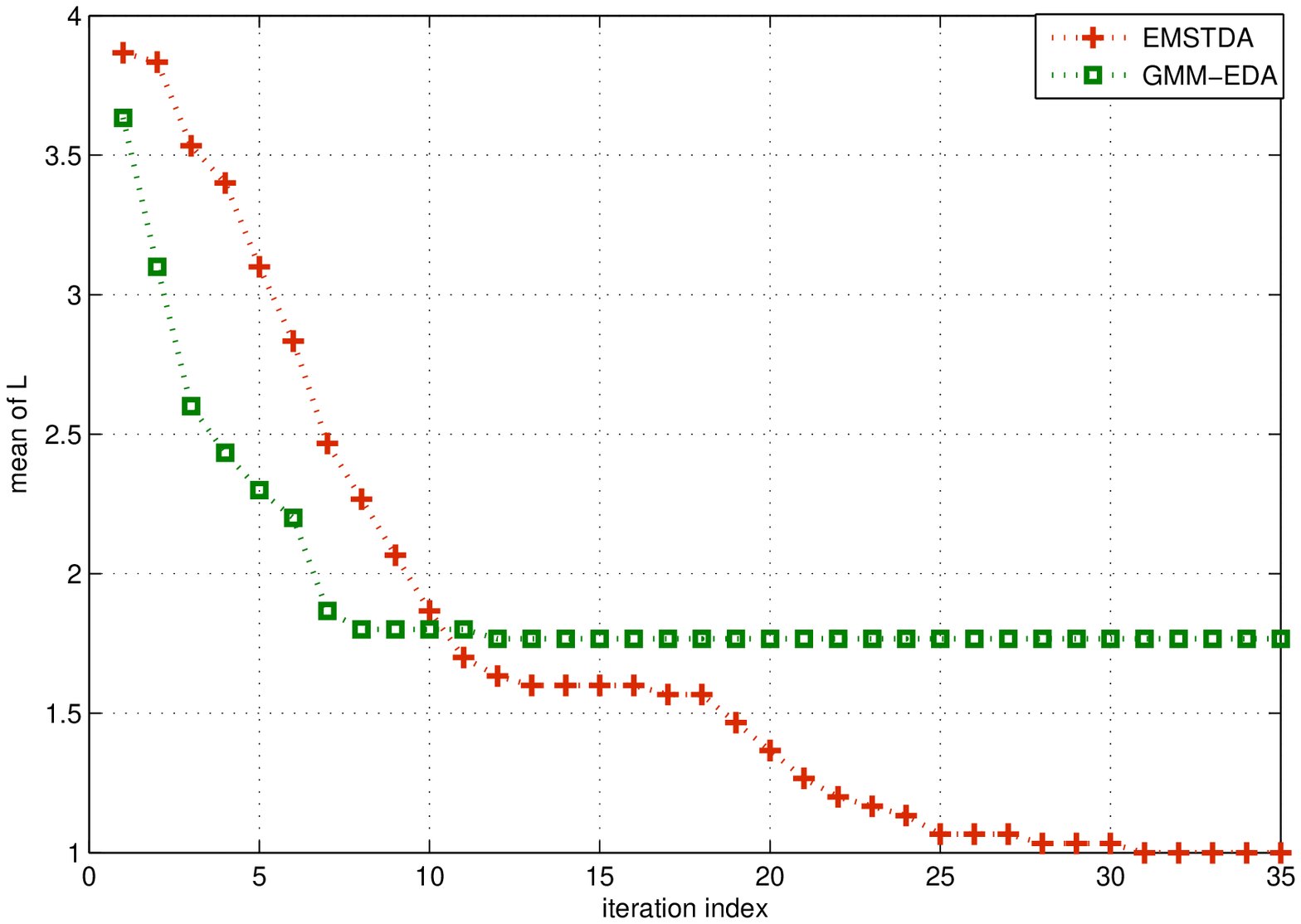}
\end{minipage}
}
\caption{Mean number of survival mixing components in the mixture model.}\label{fig:L_compare}
\end{figure}

To investigate the effect of the component deletion operator used in Algorithms \ref{algo:EM_t_mixture} and \ref{algo:EM_GMM}, we record the number of
survival mixing components per iteration of the ESTDA and EMSTDA, and calculate its mean averaged over 30 independent runs of each algorithm. The result
for six typical objective functions is depicted in Fig. \ref{fig:L_compare}. We see that, for both algorithms, the mean number of survival components
decreases as the algorithm iterates on. This phenomenon suggests that the mixture model mainly takes effect during the early exploration stage, while,
when the algorithm goes into the latter exploitation phase, there usually remains much less components in the mixture model.
\begin{table*}\centering\small
\caption{Convergence results yielded from 30 independent runs of each algorithm on each test problem. $a\pm b$ in the table denotes that
the average and standard error of the best fitness values obtained from 30 independent runs of the corresponding algorithm are $a$ and $b$,
respectively. The best solution for each problem is marked with bold font.}
\begin{tabular}{llccccc}
\hline %
\multicolumn{2}{c}{Test Problems}& Goal:$f(x^{\star})$ &ESTDA&EMSTDA&Gaussian-EDA&GMM-EDA\\\hline
Ackley &2D& 0 &\textbf{0$\pm0$}  &0.0128$\pm0.0701$ &3.1815$\pm1.3004$  &1.5332$\pm1.8186$ \\\hline
Dejong N.5 &2D&1 & 18.1207$\pm$1.8130  & \textbf{3.2370}$\pm$2.6410 &19.6536$\pm$1.1683 &6.4677$\pm$6.3530 \\\hline
Easom &2D&-1 &-0.9330$\pm$0.2536  &\textbf{-0.9587}$\pm$0.1862 &0.2262$\pm$0.4186 &-0.3153$\pm$0.4589 \\\hline
Rastrigin &2D&0 &\textbf{0}$\pm0$  &0.0050$\pm0.0202$ &\textbf{0}$\pm$0 &0.0182$\pm0.0643$ \\\hline
    &5D&0 &\textbf{0}$\pm2.13\times10^{-12}$ &0.6562$\pm0.7985$&0$\pm1.70\times10^{-11}$&0.3575$\pm0.6293$\\\cline{2-7}
    &10D&0 &\textbf{0.0383}$\pm0.0447$&0.5383$\pm0.5980$&0.0396$\pm0.0272$&0.5341$\pm0.8317$\\\hline
Michalewicz &2D&-1.8013 &\textbf{-1.8013}$\pm0$  &\textbf{-1.8013}$\pm0$ &\textbf{-1.8013}$\pm0$&\textbf{-1.8013}$\pm0$ \\\hline
    &5D&-4.6877 &\textbf{-4.6877}$\pm9.36\times10^{-9}$&-4.6404$\pm0.0561$&-4.6500$\pm0.0099$&-4.6459$\pm0.0175$\\\cline{2-7}
    &10D&-9.66015&\textbf{-9.5384}$\pm0.0475$ &-9.4226$\pm0.2107$&-9.1047$\pm0.1353$&-9.1426$\pm0.1415$\\\hline
L\`{e}vy N.13 &2D&0 &\textbf{0}$\pm0$  &0.0014$\pm0.0053$ &\textbf{0}$\pm$0&0.0043$\pm$0.0179 \\\hline
Cross-in-tray &2D&-2.0626&\textbf{-2.0626}$\pm$0&\textbf{-2.0626}$\pm$0&\textbf{-2.0626}$\pm$0&\textbf{-2.0626}$\pm$0\\\hline
Drop-wave &2D&-1&-0.9884$\pm0.0129$&-0.9909$\pm0.0125$ &\textbf{-0.9990}$\pm0.0010$&-0.9938$\pm0.0097$ \\\hline
Eggholder &2D&-959.6407&-588.9196$\pm$75.2446&\textbf{-731.8013}$\pm$145.5383&-560.7251$\pm$4.2080&-686.5236$\pm$147.5427  \\\hline
Griewank &2D&0&19.5764$\pm$3.7905&\textbf{1.5197}$\pm$5.5576&30.4232$\pm$1.0396 &15.7949$\pm$16.3001\\\hline
Holder table &2D&-19.2085&-19.0835$\pm$0.1918&\textbf{-19.2085}$\pm0$ &-19.1860$\pm$0.0821 &\textbf{-19.2085}$\pm$0 \\\hline
L\`{e}vy &2D&0&\textbf{0}$\pm$0&\textbf{0}$\pm$0 &\textbf{0}$\pm$0 &0$\pm1.0605\times10^{-4}$   \\\hline
Schaffer &2D&0&0$\pm1.7064\times10^{-6}$&0.0001$\pm4.5280\times10^{-4}$ &\textbf{0}$\pm0$ &0$\pm2.0208\times10^{-4}$   \\\hline
Schwefel &2D&0&368.3134$\pm$75.4837&\textbf{184.2835}$\pm$118.4596&436.8232$\pm$2.5521 &247.0598$\pm$123.2987  \\\hline
Shubert &2D&-186.7309&-186.7309$\pm4.05\times10^{-13}$&\textbf{-186.7309}$\pm1.64\times10^{-13}$ &-186.7309$\pm1.38\times10^{-4}$&-186.7309$\pm2.48\times10^{-5}$   \\\hline
Perm &2D&0&\textbf{0}$\pm$0 &\textbf{0}$\pm$0 &\textbf{0}$\pm$0 & 0$\pm4.0029\times10^{-6}$   \\\hline
Rosenbrock &2D&0&0.0420$\pm0.0418$&\textbf{0.0036}$\pm0.0129$&0.0477$\pm0.0558$&0.0151$\pm0.0382$ \\\hline
\end{tabular}
\label{Table:convergence values}
\end{table*}

\begin{table}[h!]
\centering
\caption{A simple statistics made for the results presented in Table \ref{Table:convergence values}.
Each figure in this table represents the number of times the corresponding algorithm have outputted a better solution than all the rest of
competitor algorithms under consideration.
Note that in some cases, e.g, the Cross-in-tray function
case, more than one algorithms provide an identical best solution. For such cases,
no algorithm will get a score in the statistics made here.}
\begin{tabular}{c|c|c|c }
\hline  ESTDA & EMSTDA & Gaussian-EDA & GMM-EDA\\
\hline  5 & 7 & 2 & 0 \\
\hline
\end{tabular}\label{Table:stat}
\end{table}

\section{Concluding remarks}
In this paper we introduced the Student's $t$ distribution, for the first time, into a generic paradigm of EDAs.
Specifically, we developed two Student's $t$ based EDAs, which we term as ESTDA and EMSTDA here.
The justification for the proposed algorithms lies in the assumption that the heavier tails of the Student's $t$ distribution
are beneficial for an EDA type algorithm to explore the solution space. Experimental results demonstrate that for most cases
under consideration, the Student's $t$ based EDAs indeed performed remarkably better than their Gaussian counterparts implemented in the same algorithm framework.
As a byproduct of our experimental study, we also demonstrated that mixture models can bring in additional advantages, for both
Student's $t$ and Gaussian based EDAs, in tackling global optimization problems.

All algorithms developed here are based on density models that use full covariance matrices. The reason is two-fold. One is for the ease of presentation of the basic idea, namely, replacing the commonly used Gaussian models by heavier tailed Student's $t$ models in EDAs. The other is for conducting a completely fair performance comparison, since, if we combine the differing state-of-art techniques into the algorithms used for comparison, it may become difficult for us to judge whether a performance gain or loss is caused by replacing Gaussian with the Student's $t$ model or the other state-of-art techniques.
Since the full covariance model based EDA framework is so basic that many state-of-art techniques can be taken as special cases of it, we argue that the ESTDA and EMSTDA algorithms presented here can be generalized by combining other techniques, such as factorizations of the covariance matrices, dimension reduction by separating out weakly dependent dimensions or random embedding, other divide and conquer or problem decomposition approaches, to handle specific optimization problems.
\section*{APPENDIX: Benchmark objective functions used in this paper}\label{sec:appendix}
Here we introduce the objective functions used in Section \ref{sec:performance_compare} for testing algorithms designed to search global minimum. Note that
in this section, we use $x_i$ to denote the $i$th dimension of $x$ for ease of exposition.
\subsection{Ackley Function}
This function is defined to be
\begin{equation}
f(x)=\left[-a\exp\left(-b\sqrt{\frac{1}{d}\sum_{i=1}^dx_i^2}\right)-\exp\left(\frac{1}{d}\sum_{i=1}^d\cos(cx_i)\right)+a+\exp(1)\right],
\end{equation}
where $a=20, b=0.2$ and $c=2\pi$. This function is characterized by a nearly flat outer region, and a large peak at the centre, thus it poses a risk for optimization algorithms to be trapped
in one of its many local minima. The global minima is $f(x^{\star})=0$, at $x^{\star}=(0,\ldots,0)$.
\subsection{De Jong Function N.5}
The fifth function of De Jong is multimodal, with very sharp drops on a mainly flat surface. The function is evaluated on the square $x_i\in[-65.536,65.536]$, for all $i=1,2$, as follows
\begin{equation}
f(x)=\left(0.002+\sum_{i=1}^25\frac{1}{i+(x_1-a_{1i})^6+(x_2-a_{2i})^6}\right)^{-1},
\end{equation}
where
\begin{eqnarray}
\mathbf{a}&=&\left(\begin{array}{ccccccccc}
-32&-16&0& 16 & 32 & -32 & 9 & 16 & 32\\
-32&-32&-32 & -32 & -32 & -16 & 32 & 32 & 32
\end{array}\right).\nonumber
\end{eqnarray}
The global minimum $f(x^{\star})\approx1$ \cite{yao1999evolutionary}.
\subsection{Easom Function}
This function has several local minimum. It is unimodal, and the global minimum has a small area relative to the search space.
It is usually evaluated on the square $x_i\in[-100,100]$, for all i = 1, 2, as follows
\begin{equation}
f(x)=\cos(x_1)\cos(x_2)\exp\left(-(x_1-\pi)^2-(x_2-\pi)^2\right).
\end{equation}
The global minimum $f(x^{\star})=-1$ is located at $x^{\star}=(\pi,\pi)$.
\subsection{Rastrigin Function}\label{sec:rastrigin}
This function is evaluated on the hypercube $x_i\in[-5.12,5.12]$, for all $i = 1,\ldots,d$, with the form
\begin{equation}
f(x)=\left(10d+\sum_{i=1}^d[x_i^2-10\cos(2\pi x_i)]\right).
\end{equation}
Its global minimum $f(x^{\star})=0$ is located at $x^{\star}=(0,\ldots,0)$.
This function is highly multimodal, but locations of the local minima are regularly distributed.
\subsection{Michalewicz Function}
This function has $d!$ local minima, and it is multimodal. The parameter $m$ defines the steepness of the valleys and ridges;
a larger $m$ leads to a more difficult search. The recommended value of $m$ is $m=10$.
It is usually evaluated on the hypercube $x_i\in[0,\pi]$, for all $i=1,\ldots,d$, as follows
\begin{equation}
f(x)=\sum_{i=1}^d\sin(x_i)\sin^{2m}\left(\frac{ix_i^2}{\pi}\right).
\end{equation}
The global minimum in 2D case is $f(x^{\star})=-1.8013$ located at $x^{\star}=(2.20,1.57)$, in 5D case is $f(x^{\star})=-4.687658$ and
in 10D case is $f(x^{\star})=-9.66015$.
\subsection{L\`{e}vy Function N.13}
This function is evaluated on the hypercube $x_i\in[-10,10]$, for $i = 1,2$, as below
\begin{equation}
f(x)=\left(\sin^2(3\pi x_1)+(x_1-1)^2[1+\sin^2(3\pi x_2)]+(x_2-1)^2[1+\sin^2(2\pi x_2)]\right),
\end{equation}
with global minimum $f(x^{\star})=0$, at $x^{\star}=(1,\ldots,1)$.
\subsection{Cross-in-tray Function}
This function is defined to be
\begin{equation}
f(x)=\left(-0.0001\left(\left|g(x_1,x_2)\right|+1\right)^{0.1}\right),
\end{equation}
where
\begin{equation}
g(x_1,x_2)=\sin(x_1)\sin(x_2)\exp\left(\left|100-\frac{\sqrt{x_1^2+x_2^2}}{\pi}\right|\right). \nonumber
\end{equation}
It is evaluated on the square $x_i\in[-10,10]$, for $i = 1,2$, with $f(x^{\star})=-2.06261$.
This function has multiple global minima located at $x^{\star}=$ $(1.3491,-1.3491),$ $(1.3491,1.3491),$ $(-1.3491,1.3491)$ and $(-1.3491,-1.3491)$.
\subsection{Drop-wave Function}
This is multimodal and highly complex function. It is evaluated on the square $x_i\in[-5.12,5.12]$, for $i = 1,2$, as follows
\begin{equation}
f(x)=-\frac{1+\cos(12\sqrt{x_1^2+x_2^2})}{0.5(x_1^2+x_2^2)+2}.
\end{equation}
The global minimum $f(x^{\star})=-1$ is located at $x^{\star}=(0,0)$.
\subsection{Eggholder Function}
This is a difficult function to optimize, because of a large number of local minima. It is evaluated on the square $x_i\in[-512,512]$, for $i = 1,2,$ as follows
\begin{equation}
f(x)=\left[-(x_2+47)\sin\left(\sqrt{\left|x_2+\frac{x_1}{2}+47\right|}\right)-x_1\sin\left(\sqrt{|x_1-(x_2+47)|}\right)\right].
\end{equation}
The global minimum $f(x^{\star})=-959.6407$ is located at $x^{\star}=(512,404.2319)$.
\subsection{Griewank Function}
This function has many widespread local minima, which are regularly distributed. It is usually evaluated on the hypercube $x_i\in[-600,600]$,
for all $i = 1,\ldots,d$, as follows
\begin{equation}
f(x)=\left(\sum_{i=1}^d\frac{x_i^2}{4000}-\prod_{i=1}^d\cos\left(\frac{x_i}{\sqrt{i}}\right)+1\right).
\end{equation}
The global minimum $f(x^{\star})=0$ is located at $x^{\star}=(0,\ldots,0)$.
\subsection{Holder table Function}
This function has many local minimum, with four global one. It is evaluated on the square $x_i\in[-10,10]$, for $i = 1,2$, as follows
\begin{equation}
f(x)=-\left|\sin(x_1)\cos(x_2)\exp\left(\left|100-\frac{\sqrt{x_1^2+x_2^2}}{\pi}\right|\right)\right|.
\end{equation}
Its global minimum $f(x^{\star})=-19.2085$ is located at $x^{\star}=(8.05502,9.66459),(8.05502,-9.66459),(-8.05502,9.66459)$ and $(-8.05502,-9.66459)$.
\subsection{L\`{e}vy Function}
This function is evaluated on the hypercube $x_i\in[-10,10]$, for all $i = 1,\ldots,d$, as follows
\begin{equation}
f(x)=\left(\sin^2(\pi w_1)+\sum_{i=1}^{d-1}(w_i-1)^2[1+10\sin^2(\pi w_i+1)]+(w_d-1)^2[1+\sin^2(2\pi w_d)]\right).
\end{equation}
where $w_i=1+\frac{x_i-1}{4}$, for all $i=1,\ldots,d$.
The global minimum $f(x^{\star})=0$ is located at $x^{\star}=(1,\ldots,1)$.
\subsection{The second Schaffer Function}
This function is usually evaluated on the square $x_i\in[-100,100]$, for all $i = 1,2$. It has a form as follows
\begin{equation}
f(x)=\left(0.5+\frac{\sin^2(x_1^2-x_2^2)-0.5}{[1+0.001(x_1^2+x_2^2)]^2}\right).
\end{equation}
Its global minimum $f(x^{\star})=0$ is located at $x^{\star}=(0,0)$.
\subsection{Schwefel Function}
This function is also complex, with many local minima. It is evaluated on the hypercube $x_i\in[-500,500]$, for all $i = 1,\ldots,d$, as follows
\begin{equation}
f(x)=\left(418.9829d-\sum_{i=1}^dx_i\sin(\sqrt{|x_i|})\right).
\end{equation}
Its global minimum $f(x^{\star})=0$ is located at $x^{\star}=(420.9687,\ldots,420.9687)$.
\subsection{Shubert Function}
This function has several local minima and many global minima. It is usually evaluated on the square $x_i\in[-10,10]$, for all $i = 1,2$.
\begin{equation}
f(x)=\left(\sum_{i=1}^5i\cos((i+1)x_1+i)\right)\left(\sum_{i=1}^5i\cos((i+1)x_2+i)\right).
\end{equation}
Its global minimum is $f(x^{\star})=-186.7309$.
\subsection{Perm Function 0,d,$\beta$}
This function is evaluated on the hypercube $x_i\in[-d,d]$, for all $i = 1,\ldots,d$, as follows
\begin{equation}
f(x)=\left(\sum_{i=1}^d\left(\sum_{j=1}^d(j+\beta)\left(x_j^i-\frac{1}{j^i}\right)\right)^2\right).
\end{equation}
Its global minimum $f(x^{\star})=0$ is located at $x^{\star}=(1,\frac{1}{2},\ldots,\frac{1}{d})$.
\subsection{Rosenbrock Function}
This function is also referred to as the Valley or Banana function. The global minimum lies in a narrow, parabolic spike.
However, even though this spike is easy to find, convergence to the minimum is difficult \cite{picheny2013benchmark}. It has the following form
\begin{equation}
f(x)=\sum_{i=1}^{d-1}[100(x_{i+1}-x_i^2)^2+(x_i-1)^2],
\end{equation}
and is evaluated on the hypercube $x_i\in[-5,10]$, for all $i = 1,\ldots,d$.
The global minimum $f(x^{\star})=0$ is located at $x^{\star}=(1,\ldots,1)$.
\bibliographystyle{IEEEtran}
\bibliography{mybibfile}
\end{document}